\documentclass[aps,prr,superscriptaddress,footinbib,twocolumn,longbibliography]{revtex4-2}

\usepackage{mathrsfs}
\usepackage{amsfonts}
\usepackage{amssymb}
\usepackage{amsmath}
\usepackage{graphicx}
\usepackage[usenames,dvipsnames]{color}
\usepackage[colorlinks=true,citecolor=blue,linkcolor=magenta]{hyperref}
\usepackage{cleveref}
\usepackage{lmodern}
\usepackage{amsthm}
\usepackage{dsfont}
\usepackage{natbib}
\usepackage{bm}
\usepackage{tabularx}

\usepackage{algorithm}
\usepackage[noend]{algpseudocode}

\newcommand{\ouralgorithm}{\texttt{FMQUBOS}}

\begin{document}

\title{
Surrogate modeling via factorization machine and Ising model with enhanced higher-order interaction learning}

\author{Anbang Wang}
\email{wanganbang@cmss.chinamobile.com}
\affiliation{Future Science and Technology Research Lab, 
China Mobile (Suzhou) Software Technology Company Limited, Suzhou 215163, China}
\affiliation{Graduate School of China Academy of Engineering Physics, Beijing 100193, China}
\author{Dunbo Cai}
\affiliation{Future Science and Technology Research Lab, 
China Mobile (Suzhou) Software Technology Company Limited, Suzhou 215163, China}
\author{Yu Zhang}
\affiliation{National Laboratory of Solid State Microstructures, School of Physics, 
Nanjing University, Nanjing 210093, China}
\author{Yangqing Huang}
\affiliation{BrightGene Bio-Medical Technology Co., Ltd, Suzhou 215000, China}
\author{Xiangyang Feng}
\affiliation{BrightGene Bio-Medical Technology Co., Ltd, Suzhou 215000, China}
\author{Zhihong Zhang}
\email{zhangzhihong@cmss.chinamobile.com}
\affiliation{Future Science and Technology Research Lab, 
China Mobile (Suzhou) Software Technology Company Limited, Suzhou 215163, China}

\begin{abstract}
Recently, a surrogate model was proposed that employs a factorization machine 
to approximate the underlying input-output mapping of the original system, 
with quantum annealing used to optimize the resulting surrogate function. 
Inspired by this approach, we propose an enhanced surrogate model 
that incorporates additional slack variables into both the factorization machine 
and its associated Ising representation, thereby unifying what was by design
a two-step process into a single, integrated step. During the training phase, 
the slack variables are iteratively updated, enabling the model to account 
for higher-order feature interactions. We apply the proposed method 
to the task of predicting drug combination effects. 
Experimental results indicate that the introduction of slack variables leads to 
a notable improvement of performance. Our algorithm offers a promising approach 
for building efficient surrogate models that exploit potential quantum advantages.
\end{abstract}

\maketitle

\section{Introduction}

Our understanding of the real world is often limited, 
as nature only reveals the inputs and outputs of what appears to be a black-box system. 
To address this limitation, surrogate models~\cite{Alexander2009RecentAdvancesInSurrogateBasedOptimization} 
are constructed using the limited available input-output data, 
serving as approximations of the underlying black-box system. Compared to directly querying the black-box system, 
surrogate models offer greater efficiency and can be used 
to predict outputs for a wide range of inputs or to identify optimal input configurations.
Popular surrogate modeling techniques include Gaussian processes~\cite{Rasmussen2005GaussianProcesses}, 
neural networks~\cite{Goodfellow2016DeepLearning}, 
polynomial regression models~\cite{Myers2016ResponseSurfaceMethodology}, 
and radial basis function models~\cite{Buhmann2003RadialBasisFunctions}. 
Each of these approaches offers different trade-offs 
in terms of accuracy, interpretability, expressibility, and computational efficiency.

Recently, Kitai {\it et al}. employed a factorization machine (FM) ~\cite{Rendle2010FactorizationMachines} 
as a surrogate model for designing metamaterials~\cite{Kitai2020DesigningMetamaterials}. 
The quadratic unconstrained binary optimization (QUBO) problem derived from the FM-based surrogate model is NP-hard, 
as is its corresponding Ising formulation. Therefore, the authors utilized 
the D-Wave quantum annealer~\cite{Johnson2011QuantumAnnealingWithManufacturedSpins} to search for optimal solutions.
Quantum annealing, which is an implementation of adiabatic quantum computing~\cite{Albash2018AdiabaticQuantumComputation}, 
has the potential to explore vast solution spaces more efficiently 
than classical optimization algorithms~\cite{King2025BeyondClassicalComputation}. 
This offers alternative opportunities for tackling complex optimization tasks that are central to many NP-hard problems.
By efficiently handling highly expressive, NP-hard surrogate models, 
adiabatic quantum computing could reshape our understanding of the traditional trade-off 
between model expressibility and computational efficiency.

The surrogate model proposed by Kitai {\it et al}. is a direct combination of an FM and an Ising model, 
which limits its ability to capture interactions to only second-order (quadratic) relationships 
between input variables. In this paper, we propose an enhanced surrogate modeling framework 
that introduces additional slack variables to enable iterative refinement and ensure compatibility 
between the FM and the Ising formulation. This approach transforms what was originally 
a two-step procedure into a unified and integrated process.
As a result, our model can better exploit potential quantum advantages 
and has the capacity to capture more complex, higher-order interactions, thereby enhancing its expressive power. 
We evaluate the performance of our surrogate model on the task of predicting drug combination effects.
The remainder of this paper is organized as follows: 
Section~\ref{sec:background} reviews the fundamentals of QUBO and FM, along with the related work. 
In Sec.~\ref{sec:qubofm}, we introduce our algorithm in detail. 
Section~\ref{sec:results} presents the numerical experiments conducted on the drug combination prediction problem. 
Finally, we conclude the paper in Sec.~\ref{sec:conclusions}.

\section{Background}
\label{sec:background}
\subsection{Quadratic unconstrained binary optimization and Ising model}

The quadratic unconstrained binary optimization problem, 
also known as the unconstrained binary quadratic programming problem, 
is a well-known NP-hard combinatorial optimization 
problem~\cite{Kochenberger2014TheUnconstrainedBinary, Glover2022QuantumBridgeAnalyticsI}. 
Given a set of binary variables $\boldsymbol{x} = \{x_1, x_2, \ldots, x_n\}$, 
where each variable takes a value in $\{0, 1\}$, 
the objective of a QUBO problem is to find the assignment of $\boldsymbol{x}$ that minimizes the following function:
\begin{align}
q(\boldsymbol{x}) &= \boldsymbol{x}^T Q \boldsymbol{x} = \sum_{i,j} Q_{ij} x_i x_j,\label{eq:qubo}
\end{align}
where $Q$ is a matrix whose entries are $Q_{ij}$. QUBO serves as a unified framework (see Appendix~\ref{app:qubo}) 
for modeling a wide range of combinatorial optimization problems, 
including, but not limited to, the Max-Cut problem, the Max-SAT problem, and the quadratic assignment problem.

The QUBO problem is mathematically equivalent to the Ising model, 
a statistical mechanics model that describes the energy of a physical system 
as a function of its spin configurations. 
Let the spin configuration of a physical system be represented by 
$\boldsymbol s = \{s_1, s_2, \ldots, s_n \}$, where each $s_i \in \{-1, 1\}$. 
The energy of the Ising model is defined as
\begin{align}
E(\boldsymbol{s}) &= \sum_{i,j} J_{ij} s_i s_j + \sum_i h_i s_i,
\end{align}
where $J_{ij}$ denotes the interaction strength between spins $i$ and $j$, 
and $h_i$ represents the external field strength at spin $i$.
By applying the variable transformation $s_i = 1-2x_i$, 
the Ising model can be converted into a QUBO problem, and vice versa. 
This equivalence allows the use of the Ising formulation to solve QUBO problems, 
and consequently, a wide range of NP-hard combinatorial optimization problems~\cite{Lucas2014IsingFormulations}.
Recently, with advances in quantum computing, 
quantum algorithms such as quantum annealing~\cite{Albash2018AdiabaticQuantumComputation,Johnson2010AScalableControl} 
and the quantum approximate optimization algorithm (QAOA)~\cite{Edward2014QuantumApproximateOptimizationAlgorithm} 
have been proposed to find the ground state of Ising models. 
These algorithms offer a promising approach for tackling combinatorial optimization problems 
and have the potential to outperform classical methods in certain settings.

If we allow spins (or binary variables) to participate in higher-order interactions, 
we obtain the higher-order unconstrained binary optimization (HUBO) problem,
which is generally more difficult to solve. 
By introducing slack variables, a HUBO problem can, in principle, be transformed into an equivalent QUBO problem, 
which can then be solved using standard QUBO solvers.

\subsection{Factorization machine}

In many machine learning tasks, we aim to model the nonlinear relationships between outputs 
and input features—or, equivalently, to capture interactions among these features. 
A straightforward approach is to introduce high-order feature interactions, 
such as $w_{ij} x_i x_j$ for features $x_i$ and $x_j$. However, this leads to a key challenge: 
The number of interaction weights $w_{ij}$ grows quadratically with the number of features.
Factorization machines are a class of supervised learning algorithms 
that address this issue by decomposing the weight matrix into two low-rank matrices~\cite{Rendle2010FactorizationMachines}. 
The mathematical formulation of a factorization machine is given by
\begin{align}
\hat{y}(\boldsymbol{x}) &= w_0 + \sum_{i=1}^n w_i x_i + \sum_{i=1}^n \sum_{j=i+1}^n w_{ij} x_i x_j \\
&= w_0 + \sum_{i=1}^n w_i x_i + \sum_{i=1}^n \sum_{j=i+1}^n
\langle {\boldsymbol v}_i, {\boldsymbol v}_j \rangle x_i x_j, \label{eq:fm}
\end{align}
where $w_0$, $\boldsymbol w = (w_1, \ldots, w_n)$ and $V = ({\boldsymbol v}_1, \ldots, {\boldsymbol v}_n)$ are real-valued parameters. 
The term $\langle {\boldsymbol v}_i, {\boldsymbol v}_j\rangle$ denotes the inner product of 
vectors ${\boldsymbol v}_i$ and ${\boldsymbol v}_j$, defined as
\begin{align}
\langle {\boldsymbol v}_i, {\boldsymbol v}_j \rangle &= \sum_{f=1}^k v_{if} v_{jf},
\end{align}
where $k$ is the dimensionality of the latent vectors. 
Typically, $k$ is chosen to be much smaller than the number of features $n$, 
which significantly reduces the total number of model parameters.
An additional advantage of FMs is that the pairwise feature interactions 
can be computed efficiently in $O(k n)$ time, rather than $O(k n^2)$~\cite{Rendle2010FactorizationMachines}. 
Thanks to their use of low-rank decomposition, 
FMs are particularly effective in applications such as recommendation systems 
and click-through rate prediction, where user-item interaction data are often highly sparse.

In some cases, we require more complex nonlinear interactions than simple quadratic ones. 
FM can be generalized to capture such higher-order feature interactions.
The higher-order factorization machine (HOFM)~\cite{Blondel2016HigherOrderFactorizationMachines} is defined as follows
\begin{align}
\hat y(\boldsymbol{x}) &= w_0 + \sum_{j=1}^n w_j x_j 
+ \sum_{j_1<j_2} \langle {\boldsymbol v}^{(2)}_{j_1}, {\boldsymbol v}^{(2)}_{j_2} \rangle x_{j_1} x_{j_2} \notag \\
&\phantom{=} + \dots + \sum_{j_1<\cdots<j_d} \langle {\boldsymbol v}^{(d)}_{j_1}, \dots, {\boldsymbol v}^{(d)}_{j_d} \rangle
x_{j_1} \dots x_{j_d}. \label{eq:hofm}
\end{align}
Here, the “inner product” for multiple vectors is defined as
\begin{align}
\langle {\boldsymbol v}^{(i)}_{j_1}, \dots, {\boldsymbol v}^{(i)}_{j_i} \rangle 
&= \sum_{f=1}^k {\boldsymbol v}^{(i)}_{j_1 f} \dots {\boldsymbol v}^{(i)}_{j_i f}.
\end{align}
In total, HOFM requires $(d-1)kn$ parameters, and its computational complexity is $O(k n d^2)$. 
While higher-order FMs offer greater expressive power, they also entail significantly increased computational costs.
Therefore, when deploying such models in real-world applications, 
it is crucial to carefully balance the trade-off between model expressiveness and available computational resources.

\subsection{Combine QUBO with FM}

In FM, features are typically encoded as binary variables using one-hot 
encoding—a representation that is equivalent to the variable format used in QUBO problems. 
This structural similarity allows FM and QUBO to be naturally combined for solving 
a wide range of black-box optimization problems, 
such as designing metamaterials ~\cite{Kitai2020DesigningMetamaterials, 
Takuya2022TowardsOptimization, Kim2022High-PerformanceTransparent, Kim2024QuantumAnnealing-aided, 
Guo2024BoostingQualityFactor, Xu2025QuantumAnnealing-assisted}.

In this framework, FM is used to model the black-box system, 
while QUBO—potentially accelerated by quantum computing—is employed 
to efficiently solve the resulting optimization problem. The main steps of the algorithm are as follows:
\begin{enumerate}
\item Generate initial samples from the black-box optimization problem and form a sample set.
\item Train an FM model based on the current sample set.
\item Extract a QUBO model from the trained FM.
\item Solve the QUBO model and add its solutions to the sample set.
\item Repeat steps 2--4 until the best solution of the QUBO converges.
\end{enumerate}
We refer to this iterative algorithm as \texttt{FMQUBO}, and present its pseudocode in Algorithm~\ref{alg:fmqubo}.

Generalizing the algorithm to the higher-order case results in \texttt{HOFMQUBO}, 
where we replace FM with HOFM
and reduce the corresponding HUBO problem 
to a QUBO problem (see Appendix~\ref{app:algorithms} for details). The workflow can be summarized as follows:
\begin{enumerate}
\item Generate initial samples from the black-box optimization problem to form a sample set.
\item Train an HOFM model based on the current sample set.
\item Extract a HUBO model from the trained HOFM.
\item Convert the HUBO model into an equivalent QUBO formulation.
\item Solve the QUBO model and add its solutions to the sample set.
\item Repeat steps 2--5 until the best QUBO solution converges.
\end{enumerate}
However, three challenges arise in this process.
First, the computational cost of training HOFM is much higher than that of FM.
Second, converting HOFM to HUBO requires an exponential number of coefficient 
evaluations in the HUBO model. Third, reducing HUBO to QUBO typically 
necessitates introducing a large number of slack variables, 
which can render the resulting optimization problem computationally intractable.
For a detailed discussion about the compuational complexity, 
see Appendix~\ref{app:complexity}.

\section{Our algorithm}
\label{sec:qubofm}

Our motivation lies in the observation that the conventional 
workflow—training an HOFM and then reducing the corresponding HUBO problem 
to a QUBO problem by introducing slack variables—can be simplified. 
Instead of following this two-step process, we propose to directly incorporate 
additional slack variables during the FM training process. 
As a result, the resulting QUBO model inherently includes these slack variables, 
eliminating the need for a separate reduction step.
In this section, we present our algorithm in detail, assuming no prior knowledge of previous work.

We are dealing with a black-box function $y=f(x)$, whose internal mechanism is unknown. 
Evaluating this function is computationally or experimentally expensive; 
therefore, we aim to minimize the number of function evaluations.
To achieve this, we build a surrogate model—an approximation of the black-box function—which 
can be used in place of the original function for optimization or analysis. 
We consider two general approaches to defining such a surrogate model 
while requiring the minimal number of function evaluations.
The first approach involves iterative construction of the surrogate model. 
Starting with a small set of samples obtained from the black-box function, 
we construct an initial surrogate model. Then, using a specific acquisition strategy, 
we select new input points at which to evaluate the black-box function. 
The surrogate model is then updated based on these newly acquired samples. 
This iterative process allows us to reduce the total number of evaluations 
compared to other query strategies, provided all surrogate models reach the same level of accuracy.
In some scenarios, however, iterative construction may not be feasible 
due to time constraints or system limitations. In such cases, 
we turn to the second approach: determining in advance the minimum number of samples 
required to build an accurate surrogate model. 
Once this number is determined, only that many evaluations are performed.
Although the required number of samples varies across different problems, 
we can study effective sampling strategies for building surrogate models 
on a representative problem and aim to generalize these strategies to other similar tasks. 
To do so, we assume access to a large dataset of function evaluations. 
For each surrogate model construction, we use only a small subset of these samples 
and analyze how the model’s performance improves as more samples are included.
If the performance improvement plateaus at a certain sample size—indicating a saturation point—then 
the surrogate model that reaches this saturation with the fewest samples is considered the most efficient.

Another important question is how to assess the performance of the surrogate model. 
In the field of black-box optimization, the performance of a surrogate model is typically evaluated 
based on the quality of the optimal solution it helps identify. 
In contrast, in many machine learning applications, performance is assessed using a test set—that is, 
by evaluating the model’s predictive accuracy across a large number of data points, 
rather than focusing solely on the optimal point.
In previous work, the authors focused on scenarios involving interactive surrogate model construction, 
where the model is iteratively refined during the optimization process. 
In this work, we shift our focus to the case of noninteractive surrogate model construction, 
where the model is built independently of the optimization procedure, 
and its performance is evaluated using a test set, much like a machine learning task.

The surrogate model used in this paper combines FM with QUBO, 
incorporating additional slack variables introduced during the training phase. 
Suppose we are given a set of samples from 
the black-box function: $\{(\boldsymbol{x}_i, y_i)\}$ for $i = 1, \dots, n$, 
where $\boldsymbol{x}_i$ is the input vector and $y_i$ is the corresponding output.
We introduce an additional set of slack variables $\boldsymbol{s}$ 
and append them to each input vector to form an extended feature vector
$\boldsymbol{z}_i = [\boldsymbol{x}_i, \boldsymbol{s}]$.
Using the extended dataset $\{(\boldsymbol{z}_i, y_i)\}$, 
we train an FM to serve as the surrogate model. Once the model is trained, 
it is transformed into an equivalent QUBO formulation. 
We then solve the QUBO problem to find its optimal (typically minimal) solution 
and extract the values of $\boldsymbol{s}$ from the solution.
These slack variable values are subsequently used to generate new 
training data $\{(\boldsymbol{z}_i, y_i)\}$, which is then used to retrain the FM. 
This iterative process continues until the surrogate model reaches the desired level of accuracy.
The pseudocode of the algorithm is provided in Appendix~\ref{app:algorithms}. Since we 
introduce additional slack variables, we call our algorithm \ouralgorithm.

Before delving into the implementation details of our algorithm on a real-world problem, 
we provide some remarks regarding the role of slack variables.
In our approach, slack variables are introduced to capture potential high-order interactions 
within the black-box function. Unlike in the \texttt{HOFMQUBO} framework, where slack variables 
are added in a rule-based manner to reduce HUBO to QUBO (see Appendix~\ref{app:qubo}), 
our method learns the relationships between the slack variables and the original input variables 
directly from the training data and through the use of a QUBO solver.
In rule-based reduction methods, we need to introduce $O((d-2)n^d)$ slack variables
and $O((d-2)n^d)$ penalty weights.
However, as shown in Eq.~\eqref{eq:hofm}, 
the coefficients in HOFM are derived via low-rank matrix decomposition 
and are not independent. Consequently, many of these slack variables become redundant.
In contrast, in our algorithm, the number of slack variables $m$ is treated as a hyperparameter. 
Their initial values can be set arbitrarily. During the training process, 
the slack variables are updated by solving the corresponding QUBO problem. 
The motivation for using an optimization-based approach 
to update the slack variables lies in the observation that, even in the rule-based setting, 
the final values of the slack variables are ultimately fixed after optimization.
Therefore, we assume the introduced slack variables, which can be the unredundant ones among $(d-2)n^d$ , 
may also be somehow fixed after optimization.
However, there is no requirement in our framework that after the surrogate model is built, 
all slack variables remain fixed in a way that each one must take a concrete value. 
Instead, what may remain fixed is the total number of slack variables 
that take nonzero values—not their specific identities or assignments.
An additional degree of freedom would be to append a distinct slack vector 
$\boldsymbol{s}_i$ to each sample $\boldsymbol{x}_i$, and solve a QUBO problem 
to update $\boldsymbol{s}_i$ individually during the FM training phase. 
While this may further refine the surrogate model, it may lead to a more complex surrogate model.
We do not explore this extension in the current work.

\section{Numerical results}
\label{sec:results}

\subsection{Drug combination therapy}

\begin{figure}[tbhp]
\centering
\includegraphics[width=\linewidth]{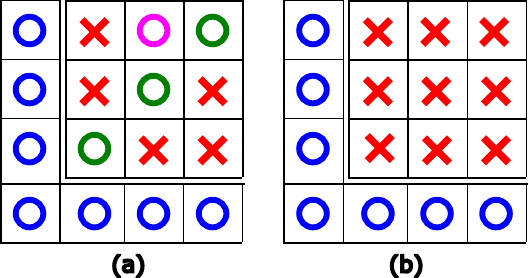}
\caption{Schematic diagram illustrating the training and test data split. 
Training data are marked by circles, and test data by crosses.
(a) Prediction of dose-response matrix for a given drug combination.
In this scenario, we select three types of data points as training samples: 
single-drug responses (shown in blue), diagonal combinations (green), 
and a set of randomly selected combination-dose pairs (magenta). 
The goal is to predict the full dose-response matrix based on these sparse observations.
(b) Prediction of unseen drug combination effects.
Here, the dataset is first divided into tested drug combinations 
and unseen drug combinations.
For the tested combinations, 
the data are further split into training and test sets following the same approach as in panel (a). 
For the unseen combinations, however, no drug combination data are available 
during training—Only the test single-drug samples (crosses) are known.}
\label{fig:scenarios}
\end{figure}
                               
Drug combination therapy~\cite{Ianevski2019PredictionDrugCombination, 
Julkunen2020LeveragingMultiwayInteractions, Paltun2021MachineLearningApproaches} 
refers to the use of two or more drugs in combination to treat complex diseases 
such as cancers and diabetes. Compared to single-drug therapies, 
combination treatments offer several advantages: 
They can enhance therapeutic efficacy through synergistic effects, 
reduce the required dosage of individual drugs to minimize side effects, 
and help overcome the development of drug resistance.
However, identifying the most effective drug combinations 
and dosages involves exploring a vast combinatorial space, 
which typically requires extensive clinical research that is costly, 
time-consuming, and sometimes infeasible. As a result, 
this problem can be viewed as a costly black-box optimization task.
In this work, we apply our algorithm to predict the effectiveness of anticancer drug combinations.

Cancer is a complex and heterogeneous disease. 
Each tumor consists of diverse cell populations that differ in their genetic mutations. 
As a result, the effect of a drug must be evaluated across various cell types, 
commonly referred to as cell lines. The outcome of a single-drug response is determined by three key factors: 
the identity of the drug, its dosage, and the specific cancer cell line being tested. 
These data are typically obtained from experimental studies.
In this work, we focus on combinations of two drugs. 
The variables involved include: drug 1, dose of drug 1, drug 2, dose of drug 2, and the cell line. 
The number of possible drug-dose-cell combinations grows exponentially 
with the number of drugs and dose levels, making comprehensive experimental evaluation impractical.
Therefore, the goal is to predict the response of these drug combinations 
based on a limited set of experimental data. According to Ref.~\cite{Julkunen2020LeveragingMultiwayInteractions}, 
there are three practical prediction scenarios in drug combination studies. 
We focus on two of them, with slight modifications:
\begin{enumerate}
   \item Prediction of the dose-response matrix for a given drug combination: 
   Given a fixed pair of drug and cell line, predict the response across different dose levels.
   \item Prediction of unseen drug combination effects for a given cell line: 
   Given a specific cancer cell line, predict the effectiveness of new drug combinations.
\end{enumerate}
These prediction tasks reflect realistic use cases and allow us to evaluate the generalization capability 
of our model under data-scarce conditions.


\subsection{Prediction of the dose-response matrix for a given drug combination}

\begin{figure}[tbhp]
\centering
\includegraphics[width=\linewidth]{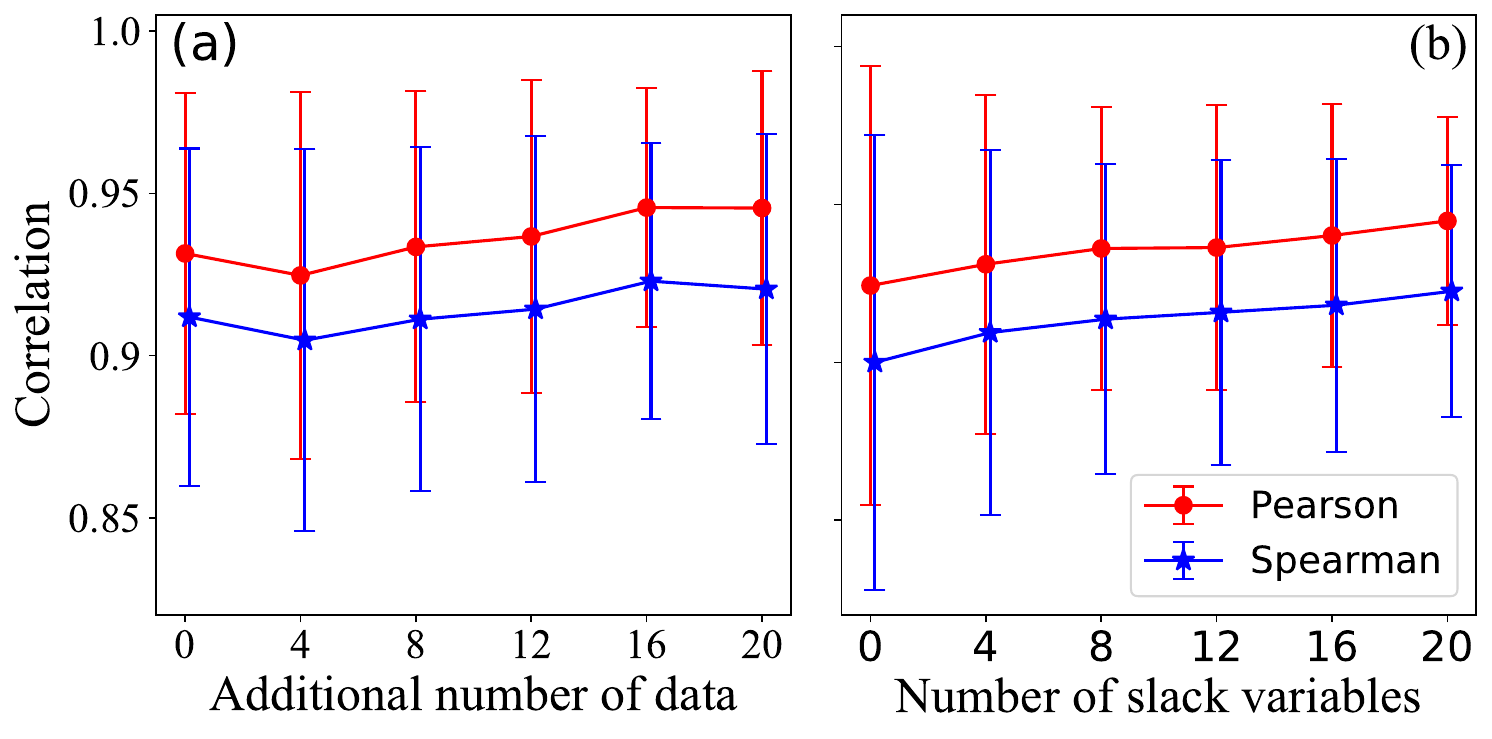}
\caption{Correlations in the first prediction scenario, 
which focuses on reconstructing the dose-response matrix for a given drug combination. 
Each matrix is trained independently, and the reported correlations 
are averaged over all $192$ drug combinations.}
\label{fig:sc-correlations}
\end{figure}

If the combination of two drugs and the cell line are known, 
the response becomes a function of the two drug doses. 
We use the dataset from Ref.~\cite{Ianevski2019PredictionDrugCombination}, 
which includes $192$ anticancer drug combinations tested across $10$ breast cancer cell lines. 
Each drug is evaluated at eight distinct concentration levels. As a result, 
for each drug-drug-cell line combination, the dose-response matrix contains $8\times 8$ entries.
We assume that some entries in the matrix are missing and aim to recover them. 
The training data consist of three components: 
the single-drug responses of the two candidate drugs, 
the diagonal combinations (i.e., equal-dose pairs after one-hot encoding), 
and a set of randomly selected off-diagonal combinations. 
The remaining entries are treated as test data. 
See Figure~\ref{fig:scenarios} for an illustration of this data-splitting scheme.
We refer to the number of randomly selected off-diagonal combinations as the additional number of training samples. 
For this dataset, each dose-response matrix is modeled independently using \ouralgorithm, 
with varying additional numbers of training data and different numbers of slack variables.
To evaluate the performance of the surrogate model, 
we compute the Pearson and Spearman correlation coefficients between 
the original (truth) and predicted matrix entries.

In Fig.~\ref{fig:sc-correlations}(a), we present the correlation 
as a function of the increasing number of additional training data points. 
The performance improves with more training data and reaches saturation 
when the number of additional training samples reaches $20$.
The correlation values shown in the figure are averaged over all $192$ drug combinations. 
In the original study~\cite{Ianevski2019PredictionDrugCombination}, 
the authors removed outliers from the training data to enhance model performance. 
However, our goal in this paper is to evaluate the effectiveness of our algorithm 
rather than to achieve state-of-the-art results on a specific task; 
therefore, we do not perform outlier detection or removal.
The impact of using slack variables is illustrated in Fig.~\ref{fig:sc-correlations}(b). 
As shown, not only does the average correlation increase with more slack variables, 
but the variance across combinations also decreases significantly. 
This indicates that the surrogate model becomes more expressive and 
stable when a greater number of slack variables is incorporated.


\subsection{Prediction of unseen drug combination effects for a given cell line}

In addition to predicting dose-response matrices for specific drug combinations, 
a more challenging task is to predict the responses of \emph{unseen drug combinations}—that is, 
combinations not observed during training. Since our algorithm is inspired by an optimization-based framework, 
we fix the cell line and aim to predict the effects of various drug combinations, 
where the variables include drug 1, its dosage, drug 2, and its dosage.
We use a subset of the NCI-ALMANAC dataset~\cite{Holbeck2017NCIALMANAC}, 
which includes $40$ drugs tested across $10$ cancer cell lines. 
The dataset contains a total of $58500$ unique drug combinations. 
From these, we select a subset as training data, 
while the remaining combinations—those never encountered during training—are used for testing.
Our goal is to evaluate whether the surrogate model can generalize to unseen drug combinations 
using only the single-drug responses as part of the training data. 
A simple illustration of this experimental setup is provided in Figure~\ref{fig:scenarios}.
We define the ratio of missing data as the proportion of unseen drug combinations 
relative to the total number of $58500$ combinations.

\begin{figure}[tbhp]
\centering
\includegraphics[width=\linewidth]{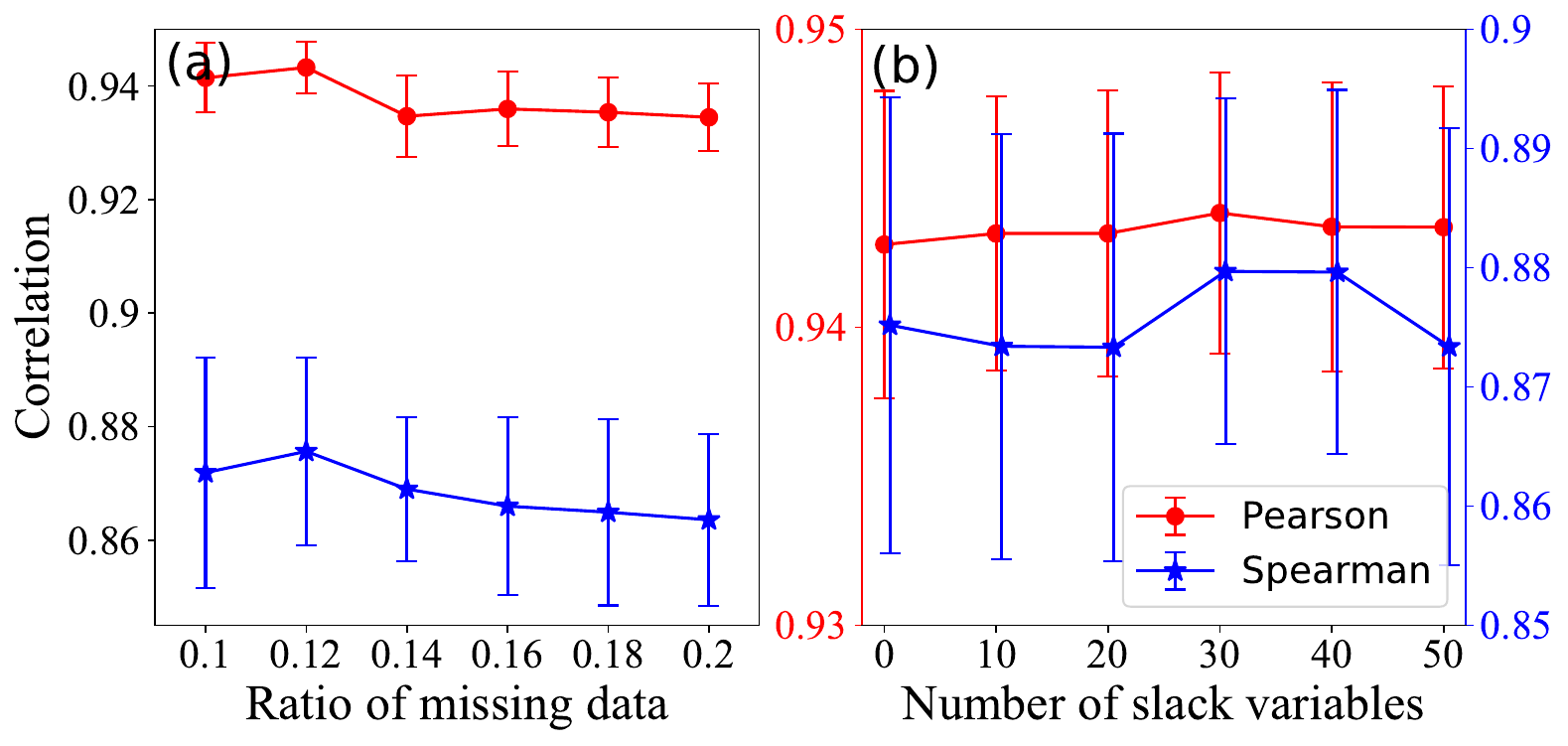}
\caption{Correlations in the second scenario on prediction of unseen drug combination effect for a certain cell line. 
The correlations are
the average of ten cell lines.}
\label{fig:c2-correlations}
\end{figure}

We conduct experiments under a similar setup as in the previous scenario. 
The results are shown in Fig.~\ref{fig:c2-correlations}. As the ratio of missing data decreases, 
the correlation increases, eventually reaching saturation when the missing data ratio drops to $0.12$.
The correlations reported in the figure are averaged over all ten cell lines. 
As the number of slack variables increases, both Pearson and Spearman correlations improve, 
while the variance across test cases decreases, which also indicates that the surrogate model becomes more expressive and stable,
consistent with the same result in the first scenario. 
However, when the number of slack variables reaches $50$, 
the Spearman correlation unexpectedly drops, whereas the Pearson correlation remains at a high level.
A possible explanation for this decline is that the model begins to overfit when too many slack variables are introduced. 
Determining an optimal number of slack variables—one that balances model expressiveness and generalization—remains 
an open issue requiring further investigation.
The differing behavior between Spearman and Pearson correlations may stem from the inherent characteristics 
of the drug combination dataset, such as nonlinear relationships or outliers that affect rank-based measures 
more strongly than linear ones.

\section{Discussions and Conclusions}
\label{sec:conclusions}

In this work, we propose a surrogate model that combines FM 
with the Ising model—or equivalently, a QUBO problem. Unlike previous approaches, 
we introduce additional slack variables into both the FM model and its corresponding Ising formulation. 
The values of these slack variables are determined during the training process using a QUBO solver.
We evaluate our algorithm on the problem of predicting drug combination effects
and demonstrate that the model incorporating slack variables 
outperforms the counterpart without slack variables 
in terms of prediction accuracy and reduced variance. 
The performance improvement arises because the slack variables 
enable the model to capture more complex interactions among input features, 
thereby enhancing its expressiveness and predictive capability.

There is an inherent trade-off between model expressiveness and computational efficiency.
Optimization of a QUBO problem—or equivalently, 
the search for the ground state of an Ising model—is known to be NP-hard. 
As a result, classical solvers are generally unable to solve such problems in polynomial time.
However, emerging quantum computing techniques—particularly quantum annealing—offer 
the potential for more efficient solutions in the near future. 
The main objective of this paper is to demonstrate a promising direction: 
As quantum technologies continue to advance, we can begin to integrate components 
that were previously considered computationally 
intractable—such as solving for the ground state of Ising models—into practical machine learning frameworks.
One such integration is the combination of FM with Ising-based surrogate models. 
This work focuses on establishing the feasibility of such hybrid modeling approaches.
Further research aimed at refining hyperparameters and optimizing model 
architecture—with the goal of achieving improved performance 
or even reaching state-of-the-art results on specific 
real-world tasks—remains an important direction for future work 
and is beyond the scope of the current study.

It should be noted that although \texttt{FMQUBOS} 
simplifies the pipeline of \texttt{HOFMQUBO} and significantly 
reduces computational complexity, its advantage relies on the assumption 
that we can efficiently solve the resulting QUBO problem—or equivalently, 
find the ground state of its corresponding Ising model. 
This problem is NP-hard. While quantum computation offers a pathway 
to solving certain instances within the complexity class BQP, 
it is widely believed—though not proven—that NP$\not\subset$BQP~\cite{Watrous2009QuantumComputationalComplexity}. 
Therefore, even with quantum computers, we are not guaranteed 
to find the ground state of an arbitrary Ising model efficiently.
However, this does not imply that all QUBO problems are intractable, 
nor that we cannot hope to develop more efficient algorithms 
than those currently available. Indeed, a rich body of research 
explores both quantum and classical approaches to tackling such problems. 
(Readers are referred to Appendix~\ref{app:complexity} 
for a discussion of the complexity of the quantum part of our algorithm.)
These include quantum algorithms such as variational quantum eigensolver (VQE) 
and QAOA~\cite{Barkoutsos2020ImprovingVariationalQuantumOptimizationUsingCVaR}, 
as well as quantum annealing~\cite{Mandra2018ADeceptiveStepTowardsQuantumSpeedupDetection,
Albash2018DemonstrationOfScalingAdvantageForQuantumAnnealerOverSimulatedAnnealing}.
While these quantum algorithms are powerful and hold long-term promise, 
they are currently limited by noise, gate fidelity, and qubit connectivity
in near-term quantum hardware. As a result, they are often restricted 
to small-scale or proof-of-principle demonstrations, 
where their performance may not yet surpass that of classical solvers. 
On one hand, these limitations may be overcome in the future with advances in quantum computing; 
on the other hand, they also motivate the exploration of alternative, 
more scalable platforms for solving Ising-type problems.
Specialized Ising machines have emerged as promising hardware solvers, 
including coherent Ising machines~\cite{Ryan2019ExperimentalInvestigation}, 
digital annealers~\cite{Aramon2019PhysicsInspiredOptimization}, 
and bifurcation machines~\cite{Goto2021HighPerformanceCombinatorialOptimizationBasedOnClassicalMechanics}. 
Alongside these, physics- or quantum-inspired classical algorithms—such as 
simulated coherent Ising machines~\cite{Reifenstein2012CoherentSATSolvers} 
and simulated quantum annealing~\cite{Albash2018DemonstrationOfScalingAdvantageForQuantumAnnealerOverSimulatedAnnealing}—have 
demonstrated strong performance.
These methods have been shown to reach or approach the best-known 
solutions on benchmark QUBO instances and, in some cases, 
exhibit speed advantages over classical optimization techniques~\cite{
   Ryan2019ExperimentalInvestigation,Aramon2019PhysicsInspiredOptimization,
   Goto2021HighPerformanceCombinatorialOptimizationBasedOnClassicalMechanics,
   Reifenstein2012CoherentSATSolvers,
   Albash2018DemonstrationOfScalingAdvantageForQuantumAnnealerOverSimulatedAnnealing,
   Zeng2024PerformanceofQAIA}.
For very large-scale optimization problems, 
it is often impossible to directly obtain the optimal solution, 
even with advanced hardware platforms. 
Hybrid methods—such as large neighborhood search~\cite{Ravindra2002SurveyOfVeryLargeScaleNeighborhoodSearchTechniques,
   Dwave2017PartitioningOptimizationProblems, Atobe2022HybridAnnealingMethod, Noguchi2023TripPlanningBasedOnSubQUBOAnnealing}
and multilevel optimization approaches~\cite{Brandt2003MultigridSolversAndMultilevelOptimizationStrategies, 
   Wang2012AMultilevelAlgorithm, Ushijima2021MultilevelCombinatorialOptimization,
   Maciejewski2024AMultilevelApproach}—have been proposed to address this challenge. 
In these frameworks, specialized hardware (e.g., quantum computers or Ising machines) 
are used to escape local minima by exploring complex energy landscapes, 
thereby increasing the likelihood of finding high-quality, 
globally optimal solutions. As a result, such hybrid approaches 
can outperform purely classical optimization algorithms.
Given these advances, we believe that surrogate models combining factorization machines 
with Ising models represent a promising direction for future research.

\section*{ACKNOWLEDGMENT}
We acknowledge the support from 
Basic Research for Application Program of China Mobile (No. R251166S).



\appendix

\section{QUBO}
\label{app:qubo}
In this section, we provide a brief introduction to QUBO and 
the process of reducing other optimization problems to QUBO.

Given a set of binary variables $\boldsymbol{x} = \{x_1, x_2, \dots, x_n\}$, 
where each variable takes values in $\{0, 1\}$, 
a QUBO problem aims to find the binary assignment that minimizes the following objective function
\begin{align}
q(\boldsymbol{x}) &=  \sum_{i<j} Q_{ij} x_i x_j + \sum_i Q_i x_i + c_0.
\end{align}
Since binary variables satisfy $x_i^2  = x_i$, the linear term $Q_i x_i$ 
can also be expressed as a quadratic term $Q_i x_i x_i$. 
By defining $Q_{ii} = Q_i$ and neglecting the constant term $c_0$, 
the QUBO problem can be rewritten in the standard form given by Eq.~\eqref{eq:qubo}.

In practice, rare problem is inherent of QUBO form. The variables may be continuous,
the interactions may be of higher order, and the problem may be with many constraints. In all cases, 
we can reduce the problems to QUBO form with the cost of either slack binary variables or
adding penalty terms to the objective function. We show how to do this below.

\emph{Nonbinary variables} Continuous or integer-valued variables can be directly encoded using binary representations. 
One common approach is to express a variable $z$ as a weighted sum of binary variables
\begin{align*}
z &= \sum_{i=-p}^{q} x_i  2^i,
\end{align*}
where $x_i\in \{0, 1\}$ are the binary variables, and $p$ and $q$ are non-negative integers 
that determine the precision and range of the representation. 
This encoding allows real or integer variables to be expressed 
in terms of binary variables suitable for QUBO or Ising formulations.

\emph{Constraints} Constraints divide the solution space into two parts: 
the feasible region, where all constraints are satisfied, 
and the infeasible region, where at least one constraint is violated. 
To ensure that the optimization variables remain within the feasible region, 
penalty terms are typically added to the objective function.
These penalty terms must satisfy the following condition: 
They are positive in the infeasible region and vanish in the feasible region. 
Let the original objective function be $q(x)$ and let the (equality) constraint be $f(x)=b$. 
Then, the modified QUBO objective function becomes
\begin{align*}
q'(x) &= q(x) + \lambda p(x) = q(x) + \lambda \left[f(x)-b\right]^2,
\end{align*}
where $p(x)$ is the penalty function and $\lambda$ is a positive weight 
that controls the strength of the constraint enforcement.

Optimization problems involving inequality constraints can also be 
transformed into QUBO form using similar techniques. 
For further details, we refer the reader to Ref.~\cite{Glover2022QuantumBridgeAnalyticsI}.

\emph{Higher-order interactions} A direct approach to handling higher-order interactions is 
to reduce them to lower-order terms ~\cite{Boros2002PseudoBoolean, 
Jiang2018QuantumAnnealingPrime, Biamonte2008Nonperturbativekbody}.
Consider the initial higher-order objective function $q(x) = x_1 x_2 x_3$. 
To reduce this to a quadratic form, we introduce a slack variable $x_4$ 
along with the constraint $x_4=x_2 x_3$. 
This results in a constrained quadratic optimization problem that minimizes
$q'(x) = x_1 x_4$ subject to $x_4=x_2 x_3$.
Adding a penalty term of the form $(x_4-x_2 x_3)^2$ does not simplify the problem, 
as it still introduces cubic terms. Instead, we proceed using the identity 
that for any binary variables $x, y, z$, the equality $xy=z$ is equivalent to
$xy+3z-2xz-2yz=0$, and otherwise, when  $xy\neq z$, the expression becomes strictly positive:
$xy+3z-2xz-2yz>0$. Using this fact, the reduced QUBO formulation becomes
\begin{align}
q''(x) &= q'(x) + \lambda p(x) \notag\\
&= x_1 x_4 + \lambda (x_2 x_3 - 3x_4 + 2x_2 x_4 + 2x_3 x_4),
\label{eq:high-order-i}
\end{align}
where $\lambda>0$ is a penalty weight. Notably, the penalty term 
$x_2 x_3 - 3x_4 + 2x_2 x_4 + 2x_3 x_4$ is always positive in the infeasible region 
where  $x_2 x_3\neq x_4$, and zero otherwise. Therefore, there is no need to square it, 
and the resulting optimization problem remains quadratic.

The reduction method described above is conceptually straightforward. 
However, for optimization problems involving many terms and higher-order interactions—such as 
those arising from HOFM—this approach requires the introduction of a large number of slack variables.
In HOFM, however, the interaction coefficients are derived through 
low-rank matrix decomposition, meaning that they are not all independent. 
As a result, the slack variables introduced during the reduction process 
may also exhibit dependencies. Therefore, it is possible that, in practice, 
fewer slack variables are actually needed than would be required 
under the assumption of full independence.

\section{Complexity analysis}
\label{app:complexity}

In this section, we present a complexity analysis of the algorithms discussed in this paper, 
including FM, HOFM, the transformation from HOFM to HUBO, 
and the subsequent conversion from HUBO to QUBO.

The straightforward computation of the interaction term in Eq.~\eqref{eq:fm} 
has a time complexity of $O(k n^2)$. Rendle~\cite{Rendle2010FactorizationMachines} 
proposed an efficient algorithm that reduces this complexity to $O(k n)$. 
He briefly mentioned an extension of FM to $d$th-order HOFM, 
claiming that the computational complexity can be reduced from $O(k n^d)$ to linear time, 
although without providing a rigorous proof. 
However, it appears that this reduction is not as straightforward as initially suggested. 
Subsequently, Blondel {\it et al}.~\cite{Blondel2016HigherOrderFactorizationMachines} 
showed that the HOFM model in Eq.~\eqref{eq:hofm} can be expressed 
using the analysis of variance (ANOVA) kernel
as
\begin{align}
\hat y(\boldsymbol{x}) &= w_0 + \sum_{j=1}^n w_j x_j 
+ \sum_{j_1<j_2} \langle {\boldsymbol v}^{(2)}_{j_1}, {\boldsymbol v}^{(2)}_{j_2} \rangle x_{j_1} x_{j_2} \notag \\
&\phantom{=} + \dots + \sum_{j_1<\dots<j_d} \langle {\boldsymbol v}^{(d)}_{j_1}, \dots, {\boldsymbol v}^{(d)}_{j_d} \rangle
x_{j_1} \dots x_{j_d} \notag \\
&= w_0 + \langle \boldsymbol w, \boldsymbol x \rangle 
+ \sum_{s=1}^{k} \mathcal A^2(\boldsymbol u_s^{(2)}, \boldsymbol x) \notag \\
&\phantom{=} + \dots + \sum_{s=1}^{k} \mathcal A^d(\boldsymbol u_s^{(d)}, \boldsymbol x),
\end{align}
where 
\begin{align}
\mathcal A^p(\boldsymbol u, \boldsymbol x) &= \sum_{j_p>\dots>j_1} 
\prod_{t=1}^p u_{j_t} x_{j_t}
\end{align}
is the homogeneous ANOVA$^p$ kernel and $\boldsymbol u_s^{(d)}$ denotes the
$s$th row vector of the matrix $V^{(d)} = ({\boldsymbol v}_1^{(d)}, \ldots, {\boldsymbol v}_n^{(d)})$.
It has been shown that the computational complexity of each ANOVA$^p$ kernel 
is $O(p n)$, leading to a total complexity 
for HOFM of $O(k(2n+\cdots +dn)) = O(k n d^2)$ for $d>2$. 
Thus, the scaling with respect to the order $d$ is quadratic—while not exponential, 
this quadratic growth may still pose computational challenges for large $d$. 
When $d=2$, the complexity becomes $O(2 k n)$, 
which is consistent with the complexity of standard FM.

The computational cost of \texttt{FMQUBO} and \texttt{HOFMQUBO} arises 
not only from the inherent complexity of FM and HOFM, 
but also from the process of converting these models into QUBO form.
The QUBO representation corresponding to FM contains $O(n^2)$ pairwise interactions, 
and each interaction coefficient must be computed directly 
as $w_{ij} = \langle \boldsymbol v_i, \boldsymbol v_j\rangle$. 
Consequently, the overall complexity of constructing a QUBO from FM is $O(k n^2)$.
Similarly, constructing a HUBO from a $d$th-order HOFM has a complexity 
of $O(k n^d)$. Furthermore, for each $d$th-order interaction term in the HUBO, 
we must introduce $d-2$ slack variables to reduce it to quadratic form. 
Hence there are in total $(d-2)n^d$ slack variables.
For each slack variable, a penalty weight $\lambda$ must be introduced, 
along with four coefficient corrections in the QUBO (or HUBO), 
as specified by Eq.~\eqref{eq:high-order-i}.
Therefore, the total complexity of transforming a $d$th-order HOFM 
into a QUBO is $O(k n^d  + 4(d-2) n^d )$, 
which simplifies to approximately $O((k+4d) n^d )$.

In addition to the exponential growth in computational complexity, 
another significant challenge is the selection of the $O(d n^d )$ penalty weights. 
These weights must be chosen carefully: They should not be too small, 
to ensure that constraints are effectively enforced, nor too large, 
to avoid overwhelming the original objective function. 
In practical applications, determining an appropriate value for 
such a large number of penalty weights can be highly nontrivial 
and poses a considerable implementation challenge.

Although the pipeline of \texttt{HOFMQUBO}—converting from HOFM to HUBO 
and then to QUBO—is conceptually straightforward, 
it is challenging to implement in practice. 
This difficulty motivates the proposal of \texttt{FMQUBOS} as a more practical alternative. 
Our method captures higher-order interactions using only FM and QUBO formulations, 
without requiring explicit modeling of higher-order terms.
The number of slack variables in \texttt{FMQUBOS} can be set to 
a moderate value—approximately half the original number of variables in our numerical example. 
As a result, the overall computational complexity of \texttt{FMQUBOS} 
is only a constant factor greater than that of \texttt{FMQUBO}. 
Moreover, \texttt{FMQUBOS} eliminates the need for manual tuning of penalty weights, 
which constitutes an additional advantage in practical deployment.

As already mentioned in Sec.~\ref{sec:conclusions}, 
the complexity of the quantum part remains an open question; 
therefore, we do not delve into detailed discussions here.
We adopt QUBO as the interface to the quantum hardware for two main reasons. 
First, evaluating higher-order interactions in HUBO 
and converting HUBO to QUBO—both discussed above—involve substantial computational costs, 
which pertain to the complexity analysis of the classical component of our algorithm.
Second, all the specialized devices we consider are capable of handling QUBO problems.
While we cannot offer a rigorous complexity analysis of the quantum part at this stage, 
we briefly comment on the three hardware platforms under consideration: 
quantum annealers, Ising machines, and universal gate-based quantum computers.

A quantum annealer is a specialized quantum computing device 
engineered to simulate the Ising model. Commercially available systems, 
such as those developed by D-Wave Systems, 
implement this paradigm using superconducting flux qubits arranged in a fixed hardware graph topology. 
Although not universal, quantum annealers offer a promising heuristic approach 
for solving hard optimization problems through adiabatic evolution. 
However, their limited connectivity poses a challenge: Many optimization 
problems—such as the one considered here—require high or even full connectivity among variables. 
Embedding such problems onto the annealer’s hardware graph is known as minor embedding, 
a task that is itself NP-hard~\cite{Lobe2024MinorEmbedding}. 
An alternative strategy is to construct a fully connected subgraph directly on the annealer, 
but this drastically reduces the number of usable qubits. 
For instance, on the D-Wave 2000Q system with $2000$ physical qubits, 
at most $65$ fully connected logical variables can be embedded~\cite{Klymko2014AdiabaticQuantumProgramming}.
In contrast, certain Ising machines—such as coherent Ising machines—are naturally fully connected, 
enabling them to tackle larger optimization problems without embedding overhead. 
That said, quantum annealers also possess unique advantages: 
They can simulate complex physical systems, such as quantum phase transitions, 
offering capabilities beyond combinatorial optimization~\cite{King2023QuantumCriticalDynamics, 
King2025BeyondClassicalComputation}. Nevertheless, these capabilities are not relevant to the problem addressed in this work.
Both VQE and QAOA operate on universal gate-based quantum computers. 
They prepare a parameterized quantum circuit, 
measure the expectation value of the problem Hamiltonian, 
and use a classical optimizer to iteratively update the parameters 
in order to solve the optimization problem. 
Although universal quantum processors may also suffer from limited qubit connectivity, 
entanglement between arbitrary qubits can be generated through carefully designed circuits. 
Moreover, universal quantum computers can directly handle HUBO problems 
without introducing auxiliary slack variables, 
thereby requiring fewer qubits for certain formulations. 
Their main drawback, however, is sensitivity to noise, 
which limits circuit depth on current hardware.
All three approaches—quantum annealing, Ising machines, 
and variational algorithms on universal devices—are heuristic in nature. 
They aim to find high-quality solutions but provide no guarantees of optimality 
or favorable scaling of computational complexity. 
Depending on the specific problem structure, 
any of these platforms may demonstrate competitive performance.

In contrast, “exact” quantum algorithms 
such as quantum phase estimation (QPE)~\cite{Nielsen2010QuantumCompuation, 
Cleve1998QuantumAlgorithmsRevisited, Abrams1999QuantumAlgorithm} 
can solve optimization problems with provable accuracy and favorable asymptotic complexity. 
However, QPE requires fault-tolerant quantum computers, which are not yet available.
Quantum imaginary time evolution (QITE) offers 
a promising middle ground~\cite{McArdle2019VariationalAnsatzBased, 
Benedetti2021HardwareEfficientVariational, Wang2025ImaginaryHamiltonianVariational}. 
It enables efficient simulation of imaginary time evolution on noisy quantum hardware 
by approximating the nonunitary dynamics with variational or local unitary operations, 
thereby facilitating the computation of ground-state energies, thermal properties, 
and solutions to combinatorial optimization problems.
Recently, a research paper~\cite{Hartung2025ConvergenceandEfficiencyProof} 
provided convergence and efficiency guarantees for QITE applied to bounded-order systems. 
Under mild assumptions, the method converges to the ground state with probability $1$ 
in a time linear in the number of qubits and inversely proportional to the energy gap. 
Furthermore, the imaginary time evolution can be compiled into a quantum circuit 
whose depth scales polynomially with the number of qubits, 
and the compilation cost itself grows only polynomially with system size and inverse gap. 
These results suggest that QITE is not only practically implementable on near-term devices 
but also comes with theoretical convergence guarantees for a broad class of relevant problems.
This may offer a concrete quantum advantage over classical methods—not merely a hypothetical or asymptotic one.
It should be emphasized, however, that these guarantees hold only 
under specific structural constraints—such as bounded interaction order—and 
do not apply universally. Consequently, this does not 
imply that $\mathsf{NP} \subseteq \mathsf{BQP}$.

\section{Pseudocode of algorithms} 
\label{app:algorithms}

In this section, we provide the implementation details of the algorithms used in this paper. 
We begin by introducing some basic concepts, notations, and functions.
We model the problem as a black-box function, denoted by \texttt{BB}, 
which takes a vector input $\boldsymbol x$ and returns a scalar output $y$ 
(although the output can be a vector in general, we consider it to be scalar for simplicity in this work). 
To extract information from the black box, we define two functions: black box sampler (\texttt{BBS}), 
which generates $n$ samples (including both inputs and corresponding outputs) from the black box and 
black box query (\texttt{BBQ}), which queries the black box with a given input $\boldsymbol x$.
Using data generated from the black box, we train surrogate models based on \texttt{FM}s
and \texttt{HOFM}s.
Although the outputs of \texttt{FM} are traditionally expressed as $w_0$, $\boldsymbol w$, and $V$ [as shown in Eq.~\eqref{eq:fm}], 
we formally pack these components into a single matrix representation denoted by $V_2$, 
where the subscript $2$ indicates that the interactions modeled by \texttt{FM} are quadratic.
For \texttt{HOFM}, an additional parameter ${d}$ is used to specify the interaction order, 
and the corresponding outputs are packed into a matrix $V_{d}$.
The output matrix $V_{d}$ is then transformed into a Hamiltonian matrix $H_{d}$ using the function \texttt{VtoH}, 
and subsequently converted into a QUBO model via the function \texttt{ReduceH}. 
Although $H_2$ already corresponds to a QUBO formulation, we still apply \texttt{ReduceH} for consistency across all orders.
The resulting QUBO model is solved using the function \texttt{QSolv}, 
which can be either a classical solver or a quantum solver—such as a quantum annealer—enabling 
us to leverage quantum acceleration if available.
The \texttt{SplitData} function is used to divide the dataset into training and test sets, 
where $n_1$ denotes the number of training samples. 
While we do not elaborate further on its internal workings, 
specific splitting strategies for each experiment are detailed in Sec.~\ref{sec:results}.
We assume access to all the aforementioned functions without concern for their internal implementations. 
A summary of these functions is provided in Table~\ref{tab:function}.
Finally, we define a composite function called \texttt{FQM} (Algorithm~\ref{alg:fqm}), which trains an \texttt{FM} model 
using training data ($X, Y$) of size $n$, and returns both the learned parameter matrix $V_2$ 
and the corresponding QUBO problem $Q$.

\begin{figure}
\begin{algorithm}[H]
\begin{algorithmic}[1]
\caption{FM-QUBO machine}
\label{alg:fqm}
\Function{FQM}{$X, Y, n, s$}
\State One-step FM QUBO machine.
\State $$ S\gets \texttt{Stack}(s, n) $$
\State $$ Z \gets [X, S] $$
\State $$ V_2 \gets \texttt{FM}(Z,Y)$$
\State $$ H_2 \gets \texttt{VtoH}(V_2)$$
\State $$ Q \gets \texttt{ReduceH}(H_2)$$
\State Output FM model and the corresponding QUBO model $[V_2, Q]$.
\EndFunction
\end{algorithmic}
\end{algorithm}
\end{figure}

In Algorithm~\ref{alg:fmqubo}, we present the pseudocode of the original algorithm 
proposed by Kitai {\it et al}. in Ref.~\cite{Kitai2020DesigningMetamaterials}. 
This algorithm is designed for black-box optimization problems, and we refer to it 
as the “FMQUBO optimization algorithm.”
The initial sample set is typically generated randomly, 
assuming no prior knowledge about the problem. In each iteration of the algorithm, 
the optimal solution predicted by the model is added to the sample set. 
The algorithm terminates when the response predicted by the QUBO model is sufficiently 
close to the true (observed) response.
Several potential improvements could be explored to enhance the performance 
of the algorithm—for instance, strategies for selecting a more informative initial sample set, 
or methods for removing less useful samples during the optimization process. 
However, such techniques are beyond the scope of this paper and will not be discussed further.

\begin{figure}
\begin{algorithm}[H]
{\small
\begin{algorithmic}[1]
\caption{{\small FMQUBO optimization algorithm}}
\label{alg:fmqubo}

\Statex
\State Input: A black box \texttt{BB} with function \texttt{BBQ} and \texttt{BBS}.
\State Input: initial sample size $n$.
\State Input: maximum number of iteration $i_{\mathrm{max}}$, tolerance $\epsilon$.
\State Get $n$ initial samples from the black box, $$[X,Y] \gets \texttt{BBS}(n)$$
\State Train a FM model with initial samples, $$V_2 \gets \texttt{FM}(X,Y)$$
\For{$i=1$ to $i_{\mathrm{max}}$}
\State $$H_2 \gets \texttt{VtoH}(V_2)$$
\State $$Q \gets \texttt{ReduceH}(H_2)$$
\State Solve the QUBO model, $$[\boldsymbol x,y] \gets \texttt{QSolv}(Q)$$.
\State Calculate the true response of $\boldsymbol x$, $$y' \gets \texttt{BBQ}(\boldsymbol x)$$
\If{$|y'-y|<\epsilon$}
\State Exit the loop.
\Else
\State Update the sample set,
$$ X\gets [X; \boldsymbol x], Y\gets [Y; y']$$
\State Train an FM model with updated sample set, 
$$V_2 \gets \texttt{FM}(X,Y)$$
\EndIf
\EndFor
\State Output: the optimal solution pair $[\boldsymbol x, y, y']$.
\end{algorithmic}
}
\end{algorithm}
\end{figure}

To capture potential {higher}-order interactions, we can replace the FM model with an HOFM, 
resulting in the HOFMQUBO optimization algorithm, as shown in Algorithm~\ref{alg:hofmqubo}.
Although the pseudocode appears similar to its FM-based counterpart, 
the \texttt{HOFM} function is significantly more complex than \texttt{FM},
and the \texttt{ReduceH} function is no longer a trivial operation. 
As a result, the implementation of \texttt{HOFMQUBO} is considerably more challenging 
compared to \texttt{FMQUBO}.

\begin{figure}
\begin{algorithm}[H]
{\small
\begin{algorithmic}[1]
\caption{{\small HOFMQUBO optimization algorithm}}
\label{alg:hofmqubo}

\Statex
\State Input: a black box \texttt{BB} with function \texttt{BBQ} and \texttt{BBS}.
\State Input: initial sample size $n$, order of FM ${d}$.
\State Input: maximum number of iterations $i_{\mathrm{max}}$, tolerance $\epsilon$.

\State Get $n$ initial samples from the black box, $$[X,Y] \gets \texttt{BBS}(n)$$
\State Train an FM model with initial samples, $$V_{d} \gets \texttt{HOFM}({d},X,Y)$$
\For{$i=1$ to $i_{\mathrm{max}}$}
\State $$H_{d} \gets \texttt{VtoH}(V_{d})$$
\State $$Q \gets \texttt{ReduceH}(H_{d})$$
\State Solve the QUBO model, $$[x,y] \gets \texttt{QSolv}(Q)$$
\State Calculate the true response of $x$, $$y' \gets \texttt{BBQ}(x)$$
\If{$|y'-y|<\epsilon$}
\State Exit the loop.
\Else
\State Update the sample set,
$$ X\gets [X, x], Y\gets [Y, y']$$
\State Train an FM model with updated sample set, 
$$V_{d} \gets \texttt{HOFM}({d},X,Y)$$
\EndIf
\EndFor
\State Output: the optimal solution pair $[x, y, y']$.
\end{algorithmic}
}
\end{algorithm}
\end{figure}

Our algorithm avoids the complexity involved in training an HOFM model 
and eliminates the need to explicitly transform HUBO problems into QUBO form.
The pseudocode for solving a black-box optimization problem using our approach 
is presented in Algorithm~\ref{alg:fmquboex}.
This algorithm follows an iterative model construction framework.

\begin{figure}
\begin{algorithm}[H]
{\small
\begin{algorithmic}[1]
\caption{{\small FMQUBOS optimization algorithm}}
\label{alg:fmquboex}

\Statex
\State Input: a black box \texttt{BB} with function \texttt{BBQ} and \texttt{BBS}.
\State Input: initial sample size $n$.
\State Input: maximum number of iterations $i_{\mathrm{max}}$, tolerance $\epsilon$.
\State Input: number of additional slack variables $m$, initial value of slack variables $\boldsymbol s$.

\State Get $n$ initial samples from the black box, $$[X,Y] \gets \texttt{BBS}(n)$$
\For{$i=1$ to $i_{\mathrm{max}}$}
\State $[V_2, Q] \gets \texttt{FQM}(X,Y,n+i-1, \boldsymbol s)$
\State Solve the QUBO model, $$[\boldsymbol z,y] \gets \texttt{QSolv}(Q)$$
\State Extract the value of slack variables from the optimal solution, 
$$\boldsymbol s \gets \boldsymbol z[-m:-1]$$
\State Extract the value of the original variables from the optimal solution, 
$$\boldsymbol x \gets \boldsymbol z[1:-m+1]$$
\State Calculate the true response of $x$, $$y' \gets \texttt{BBQ}(x)$$
\If{$|y'-y|<\epsilon$}
\State Exit the loop.
\Else
\State Update the sample set,
$$ X\gets [X; \boldsymbol x],\quad Y\gets [Y; y']$$
\EndIf
\EndFor
\State Output: the optimal solution pair $[\boldsymbol x, y, y']$.
\end{algorithmic}
}
\end{algorithm}
\end{figure}

The noniterative model construction can be formulated as a standard regression problem in machine learning.
In principle, techniques such as grid search are required to identify the optimal hyperparameters for the model.
However, we do not delve into these details in this paper.
We present the pseudocode of the FMQUBOS regression algorithm in Algorithm~\ref{alg:fmquboex2}, 
where both training and test data are provided as input.
To investigate the impact of training data size and the number of slack variables,
we run Algorithm~\ref{alg:fmquboex2} multiple times using different parameter settings.
The corresponding pseudocode for this multirun experimental setup is shown in Algorithm~\ref{alg:fmquboex3}.
The numerical results reported in this paper are based on Algorithm~\ref{alg:fmquboex3},
with implementation details provided in Sec.~\ref{sec:results}. 

In fact, both Algorithm~\ref{alg:fmqubo}, proposed by Kitai {\it et al}., 
and the algorithms introduced in this paper—including 
Algorithms~\labelcref{alg:hofmqubo,alg:fmquboex,alg:fmquboex2,alg:fmquboex3}—can be 
applied depending on the specific problem at hand and the objectives the user aims to achieve.

\section{Details of numerical calculations}
\label{app:details}
In this Appendix, we provide the details of the numerical calculations.

\subsection{Dataset}

The dataset used in the first scenario is taken from Ref.~\cite{Ianevski2019PredictionDrugCombination}. 
It contains $192$ anticancer drug combinations tested across $10$ breast cancer cell lines. 
Each drug is evaluated at eight distinct concentration levels. 
For each drug-drug-cell line combination, the dose-response matrix consists of $8\times 8$ entries.
Each entry represents the relative inhibition of the drug combination on the corresponding cell line, 
and is therefore always non-negative. In the original paper, 
the authors exploit this property by applying non-negative matrix factorization 
to predict the full dose-response matrix.
In contrast, FM do not inherently enforce non-negativity in their output, 
and we do not impose any constraints or modifications to ensure this property in our approach.

The dataset used in the second scenario is derived from the NCI-ALMANAC database. 
We use a subset provided by Ref.~\cite{Julkunen2020LeveragingMultiwayInteractions}, 
which includes $40$ selected drugs and $10$ cell lines, resulting in a total of $58500$ drug-drug-cell line combinations. 
In this dataset, each drug is tested at four distinct concentration levels.
The response variable represents the percentage growth of the treated cell line relative to the control. 
This value must be greater than $-100\%$, as negative percentages indicate growth inhibition. 
In our subset, the response values range from $-95.17\%$ to $164.18\%$.

The NCI-ALMANAC dataset is available in Ref.~\cite{NCI-ALMANAC}. The original data of the first 
scenario can be found in Ref.~\cite{DECREASE}. The original data of 
the second scenario can be found in Ref.~\cite{comboFM}. The processed data
of these two scenarios used in the paper are provided in Ref.~\cite{fmqubodata}.

\begin{figure}
\begin{algorithm}[H]
{\small
\begin{algorithmic}[1]
\caption{{\small FMQUBOS regression algorithm}}
\label{alg:fmquboex2}

\Function{FQEX}{$X_1, Y_1, X_2, Y_2, i_{\mathrm{max}}, \epsilon, m, \boldsymbol s$}
\State Input: training dataset ${X_1, Y_1}$ with size $n_1$, testing dataset ${X_2, Y_2}$ with size $n_2$.
\State Input: maximum number of iterations $i_{\mathrm{max}}$, tolerance $\epsilon$.
\State Input: number of additional slack variables $m$, initial value of slack variables $\boldsymbol s$.

\For{$i=1$ to $i_{\mathrm{max}}$}
\State $$ [V_2, Q] \gets \texttt{FQM}(X_1, Y_1, n_1, s)$$
\State $$ [z, y] \gets \texttt{QSolv}(Q)$$
\State $$s \gets z[-m:-1]$$
\State $$Y_1'\gets V_2(X_1)$$
\If{$\texttt{Loss}(Y, Y')<\epsilon$}
\State Exit the loop.
\EndIf
\EndFor
\State Output: $\texttt{Loss}(Y, Y')$.
\EndFunction
\end{algorithmic}
}
\end{algorithm}
\end{figure}

\begin{figure}
\begin{algorithm}[H]
{\small
\begin{algorithmic}[1]
\caption{{\small Testing FMQUBOS regression algorithm}}
\label{alg:fmquboex3}

\Statex
\State Input: dataset $X$, $Y$ with size $n$.
\State Input: maximum number of iterations $i_{\mathrm{max}}$, tolerance $\epsilon$.
\State Input: bounds of additional slack variables $m_a$, $m_b$ 
\State Input: bounds of training samples $n_a$, $n_b$

\State $S \gets \{\}$
\For{$n_1$ in $n_a$ to $n_b$}
\State $$[X_1, Y_1, X_2, Y_2] = \texttt{SplitData}(X, Y, n_1)$$
\For{$m$ in $m_a$ to $m_b$}
\State $$\boldsymbol s \gets [0]\times m$$
\State $$\eta \gets \texttt{FQEX}(X_1, Y_1, X_2, Y_2, i_{\mathrm{max}}, \epsilon, m, \boldsymbol s) $$
\State Add $(n_1, m, \eta)$ to $S$
\EndFor
\EndFor
\State Output: S.

\end{algorithmic}
}
\end{algorithm}
\end{figure}

\subsection{Data representation}

In the dataset, drugs and cell lines are represented as string variables, 
while drug concentrations are given as real numbers.
To represent these variables for training an FM and subsequently solving a QUBO problem, 
we employ one-hot encoding.
In the first scenario, each concentration level is encoded as an $8$-bit binary vector. 
The input vector is formed by concatenating the two encoded concentration vectors 
corresponding to the two drugs in the combination.
In the second scenario, each drug is encoded using a $40$-bit binary 
vector (representing one-hot encoding over $40$ drugs), and each concentration is encoded as a $4$-bit binary vector. 
The final input vector is constructed by concatenating two drug vectors and two corresponding concentration vectors.
Although the order of the two drugs in a combination does not affect the true biological response, 
it may influence the model’s predictions due to how features are structured in the FM. 
To mitigate this asymmetry, we duplicate the dataset and reverse the order of drugs 
and their associated concentrations in each pair during training.

\subsection{Training setup}

After one-hot encoding, the dataset is represented as a collection of pairs $\{(\boldsymbol x_i, y_i)\}$, 
where $\boldsymbol x_i$ is a $16$-bit binary vector in the first scenario 
and an $88$-bit binary vector in the second scenario, and $y_i$ is a real-valued response.
The mathematical formulation of the FM model—as also given in Eq.~\eqref{eq:fm}—is
\begin{align*}
\hat y(\boldsymbol{x}) &=
w_0 + \sum_{i=1}^n w_i x_i + \sum_{i=1}^n \sum_{j=i+1}^n
\langle {\boldsymbol v}_i, {\boldsymbol v}_j \rangle x_i x_j,
\end{align*}
where $n$ is the number of input features.
In our experiments, we set the dimension of each latent vector ${\boldsymbol v}_i$ to $4$ 
in the first scenario and to $8$ in the second scenario. 
These values are chosen to be approximately twice the number of the original features before encoding.

The objective function of the FM is defined as the mean squared error 
between the true response $y_i$ and the predicted value $\hat y(\boldsymbol x_i)$.
To improve generalization and prevent overfitting, we incorporate both $L1$-norm 
and $L2$-norm regularization terms into the objective function. 
The regularized loss function is given by
\begin{align*}
\mathcal{L} &= \frac{1}{n} \sum_{i=1}^n (y_i - \hat y(\boldsymbol x_i))^2 
+ \beta_1 |\boldsymbol w|_1 + \beta_2 |\boldsymbol v|_{\mathrm{F}}^2,
\end{align*}
where
\begin{itemize}
   \item $\boldsymbol w$ is the weight vector of the linear term in FM,
   \item $|\bullet|$ denotes the $L1$-norm, and
   \item $|\bullet|_F$ denotes the Frobenius norm (matrix extension of the $L2$-norm).
\end{itemize}
In our experiments, we set $\beta_1 = 0.02$ and $\beta_2 = 0.003$ in the first scenario
and $\beta_1 = 0.015$ and $\beta_2 = 0.002$ in the second scenario.
In both scenarios, the learning rate is fixed at $0.003$.

Since the D-Wave quantum annealer cloud service is not accessible to users in the authors’ region, 
we employ the simulated annealing algorithm provided by the D-Wave Ocean SDK to solve the QUBO models.
The QUBO solutions are obtained under the constraint that the input vector must satisfy the one-hot encoding. 
For instance, in the first scenario, the input vector $\boldsymbol x$ is a $16$-bit binary vector, 
where the first $8$ bits represent the concentration level of the first drug 
and the last $8$ bits correspond to the second drug. 
To enforce one-hot encoding, we impose the constraints
$\sum_{i=1}^{8} x_i = 1$ and $\sum_{i=9}^{16} x_i = 1$.
For each QUBO problem, the simulated annealing algorithm is executed $5000$ times, 
and the solution with the lowest energy is selected as the final result.

\subsection{Supplementary numercial results}

\begin{figure}[tbhp]
\centering
\includegraphics[width=\linewidth]{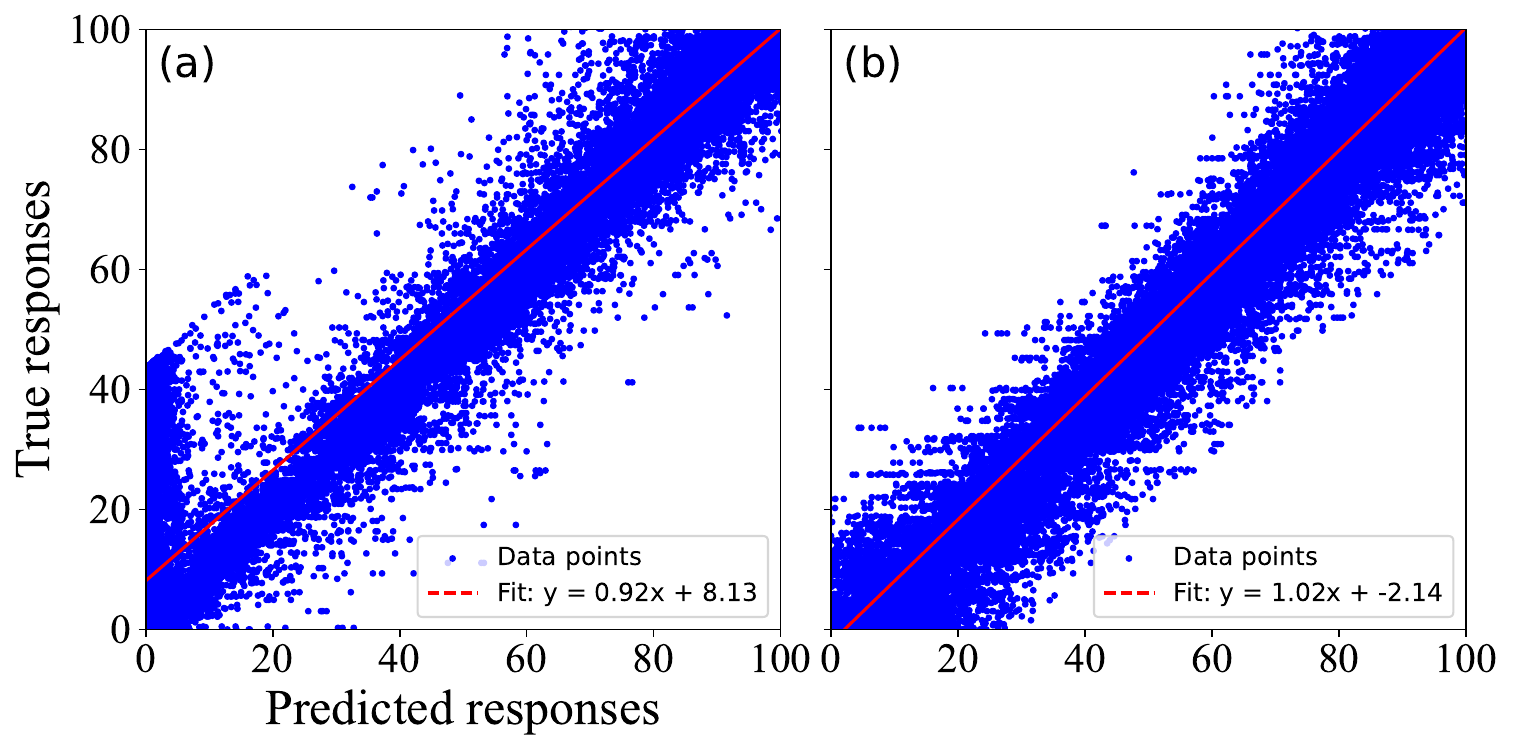}
\caption{Comparison between the true responses and the predicted responses. 
(a) No slack variable. (b) Sixteen slack variables.}
\label{fig:sc-all-resp}
\end{figure}

\begin{figure}[tbhp]
\centering
\includegraphics[width=\linewidth]{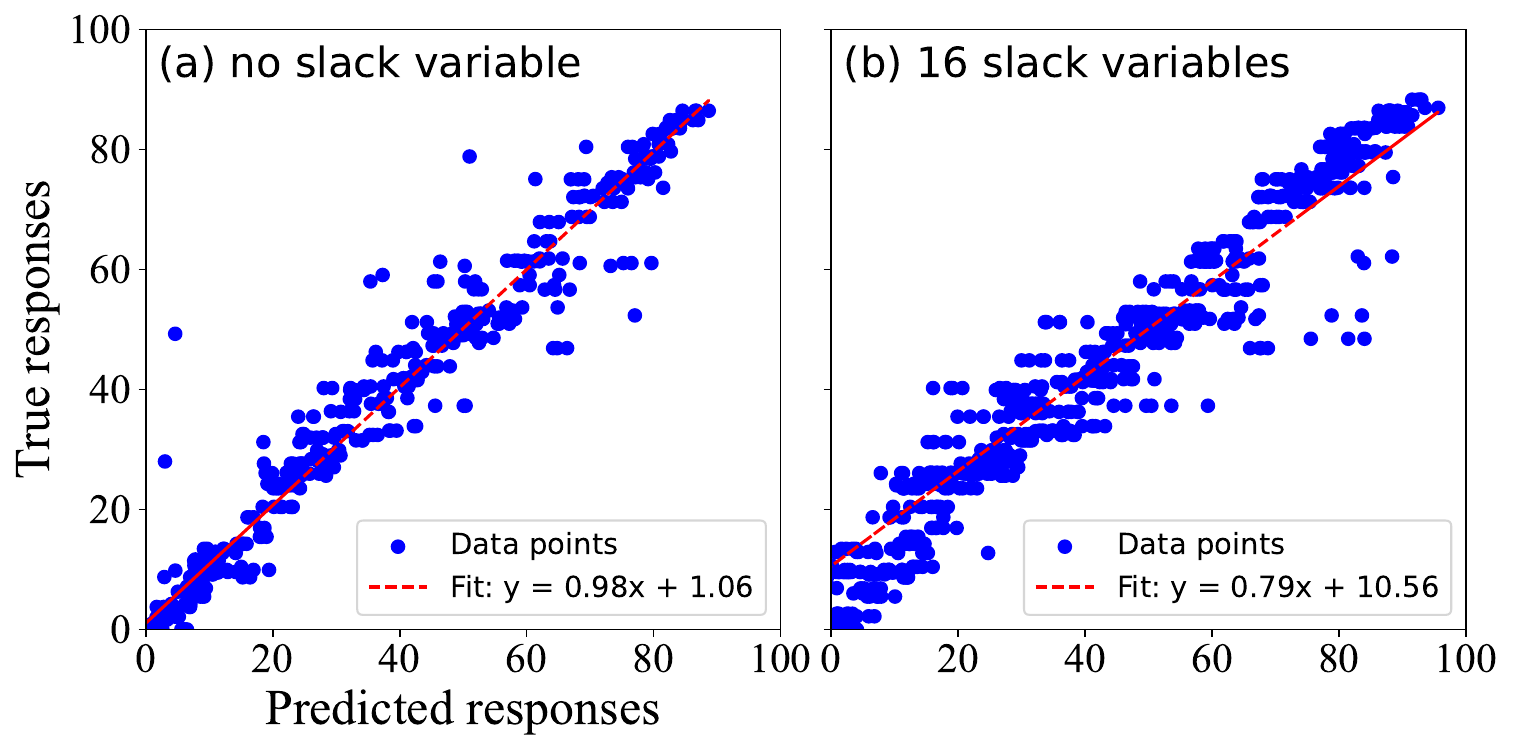}
\caption{Comparison between the true responses and the predicted responses with only 
the best responses of all the dose-response matrices are shown.
(a) No slack variable. (b) Sixteen slack variables.}
\label{fig:sc-opt-resp}
\end{figure}

In Sec.~\ref{sec:results}, we presented the correlation results obtained 
using different numbers of slack variables. 
We found that incorporating slack variables improves the algorithm’s performance, 
particularly by reducing the variance in correlation values across different runs.
To further demonstrate this effect, we present both the true responses 
and the predicted responses generated by \ouralgorithm.

In Fig.~\ref{fig:sc-all-resp}, we compare the true responses with the predicted responses in the first scenario. 
The results include all $64$ matrix entries from $192$ drug combinations across $6$ different settings of 
additional training data, excluding only those cases where the training process failed to converge.
Out of a total of $73728$ data points, Fig.~\ref{fig:sc-all-resp}(a) includes $44236$ points 
obtained without using any slack variables, while Fig.~\ref{fig:sc-all-resp}(b) 
contains $72253$ points when $16$ slack variables are used.
The primary reason for nonconvergence appears to be the presence 
of a “tail” along the y axis in Fig.~\ref{fig:sc-all-resp}(a), 
which indicates poor model predictions for certain samples. 
The surrogate model incorporating slack variables significantly reduces the emergence of this tail, 
thereby improving overall performance.
A similar comparison is presented in Fig.~2 of Ref.~\cite{Julkunen2020LeveragingMultiwayInteractions}, 
where comboFM-5 (a model based on a fifth-order FM) outperforms comboFM-2 
by producing fewer outliers in the tail region. In our work, adding slack variables achieves a comparable effect, 
enhancing prediction accuracy and stability. 
This improvement is particularly significant when the dataset is large, 
as larger datasets are more likely to contain numerous outliers—due to experimental errors, 
recording mistakes, or other data collection issues. In clean datasets without outliers, 
quadratic interactions may be sufficient to capture the underlying patterns. 
However, in the presence of outliers, higher-order interactions become necessary 
to effectively compensate for their distorting effects and maintain model robustness.

We also evaluate the prediction of the best response for each combination, 
defined as the optimal response among the $64$ matrix entries.
Fig.~\ref{fig:sc-opt-resp}(a) shows only $622$ valid results 
out of a total of $1152$ combinations when no slack variables are used.
In contrast, Fig.~\ref{fig:sc-opt-resp}(b) includes $1082$ valid points 
when $16$ slack variables are incorporated.

{
However, including more “well-predicted” responses appears to degrade the quality of the fitted line. 
While the introduction of slack variables allows our model to handle outliers responsible 
for the “long-tail” phenomenon observed in Fig.~\ref{fig:sc-all-resp}, 
this long-tail structure is not entirely eliminated when predicting the best responses. 
This may explain why the slope of the fitted line in Fig.~\ref{fig:sc-opt-resp}(b) is smaller than expected.
Importantly, this does not imply that the model with slack variables is inferior. 
On the contrary, it demonstrates improved predictive capacity: 
In Fig.~\ref{fig:sc-opt-resp}(a), only about half of the predicted responses are included in the analysis, 
whereas in Fig.~\ref{fig:sc-opt-resp}(b), $94\%$ of the predicted responses are retained. 
Despite the flatter slope, the model with slack variables successfully identifies significantly 
more high-quality responses compared to the model without slack 
variables—indicating enhanced coverage and robustness in capturing optimal outcomes.
Addressing the slight underestimation in the slope remains a potential avenue for future model enhancement.
}

\begin{figure}[htbp]
\centering
\includegraphics[width=\linewidth]{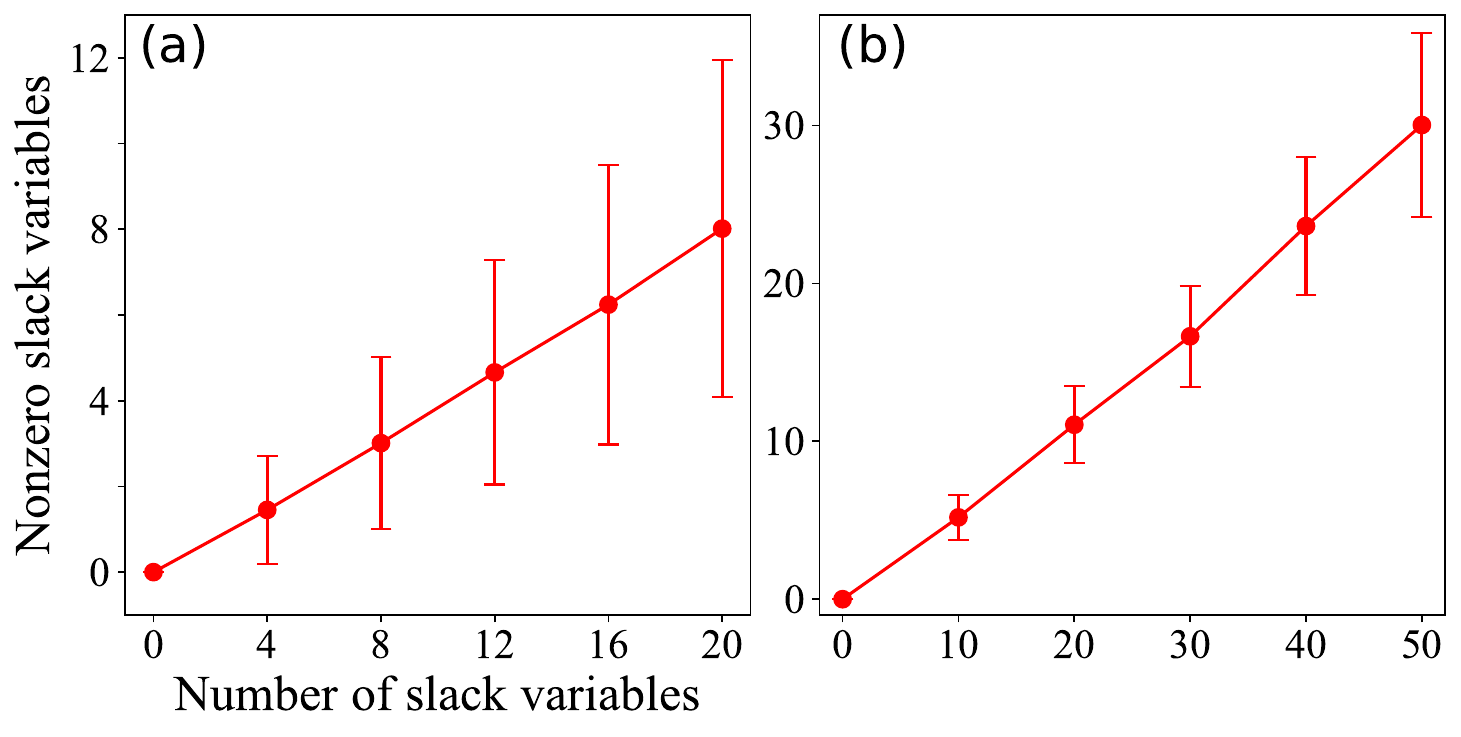}
\caption{Number of nonzero slack variables vs the number of additional slack variables in the model.}
\label{fig:nonzeros}
\end{figure}

{
In the FM model, all variables are pairwise connected. 
For each pair of variables $x_i$ and $x_j$, their interaction is defined as $w_{ij} x_i x_j$, 
which captures the strength of the relationship between the corresponding features.
The slack variables introduced in our algorithm do not depend explicitly 
on the input data but serve to encode higher-order interaction patterns. 
During training, these slack variables take on specific values, 
which can be interpreted as indicating the presence 
or absence of particular higher-order interactions.
It is essential that not all slack variables are zero; 
otherwise, the model would lack expressive capacity 
and be unable to learn any higher-order interactions. 
A larger number of nonzero slack variables enables the model 
to capture more complex interaction structures, thereby increasing its expressiveness. 
However, it is undesirable for all slack variables to become active (i.e., take value $1$), 
as this would introduce excessive model complexity, increase the risk of overfitting, 
and lead to poor generalization on unseen datasets or other tasks.
Therefore, a good trade-off between model expressiveness and generalization 
is achieved when approximately half of the slack variables are active. 
Ideally, the model should reach this balance automatically, without requiring manual tuning. 
As shown in Fig.~\ref{fig:nonzeros}, after training, 
roughly half of the slack variables take the value $1$ (i.e., are nonzero), 
indicating that the model self-regularizes 
to maintain an effective balance between flexibility and generalization.
}

\begin{table*}[htbp]
\centering
\begin{tabularx}{\linewidth}{ 
   XX} 
\hline\hline
Function & Description \\ 
\hline
$[X,Y]=\texttt{BBS}(n)$ & Generate $n$ samples from the black box, where $X$ and $Y$ are stacked inputs and outputs of the black box. \\
$y = \texttt{BBQ}(x)$ & Query the black box with input $x$. \\
$V_2 = \texttt{FM}(X,Y)$ & Implement a factorization machine.\cite{footnote1}\\
$V_d = \texttt{HOFM}({d},X,Y)$ & Implement a {higher}-order factorization machine of order ${d}$.\\
$H_{d} = \texttt{VtoH}(V_{d})$ & Transform FM output to a Hamiltonian.\\
$Q = \texttt{ReduceH}(H_{d})$ & Reduce a ({higher}-order) Hamiltonian to a QUBO model.\\
$[x,y] = \texttt{QSolv}(Q)$ & Solve a QUBO model and return the solution.\\
$S = \texttt{Stack}(s, n) $ & Stack a row vector $s$ $n$ times vertically.\\
$\eta = \texttt{Loss}(Y, Y')$ & Compute the loss function. \\
$[X_1, Y_1, X_2, Y_2]=\texttt{SplitData}(X, Y, n_1)$ & Split data into training (size $n_1$) and test sets according to some rule.\\
\hline\hline
\end{tabularx}
\caption{Basic functions used to construct algorithms.}
\label{tab:function}
\end{table*}

\bibliography{reference}

\begin{thebibliography}{60}
\bibitem [{foo()}]{footnote1}%
  \BibitemOpen
  \href@noop {} {}\bibinfo {note} {We use the same notation $V_2$ for the
  output FM model itself and the coefficients, e.g., $Y'=V_2(X)$ for the model
  and $H_2 = \texttt{VtoH}(V_2)$ for the coefficients. $V_k$ follows the same
  rule.}\BibitemShut {Stop}%
\end{thebibliography}

\end{document}